\documentclass[letterpaper,10 pt,conference]{ieeeconf}
\IEEEoverridecommandlockouts

\usepackage{amsmath,amssymb,amsfonts}
\usepackage{mathrsfs}
\usepackage{bm}
\usepackage[hidelinks]{hyperref}
\usepackage{subcaption}
\usepackage{booktabs}       
\usepackage{placeins}
\usepackage[nocompress]{cite}
\usepackage{graphicx}
\usepackage[acronym,shortcuts]{glossaries}
\usepackage{comment}
\usepackage{enumerate}
\usepackage[capitalise]{cleveref}
\usepackage[table,xcdraw,dvipsnames]{xcolor}
\usepackage{siunitx}
\usepackage{bm} 
\usepackage{amsmath,amssymb,amsfonts}
\usepackage[english]{babel}
\usepackage{microtype}
\usepackage{multirow}
\usepackage{pifont}
\usepackage{algorithm}
\usepackage[noend]{algpseudocode}
\usepackage{cuted}
\usepackage{cases}
\usepackage{textcomp}
\usepackage{xcolor}
\usepackage{fancyhdr}
\usepackage{tikz}
\usepackage{eso-pic}
\let\labelindent\relax 
\usepackage{enumitem}
\setlist[itemize]{align=parleft,left=1pt..1em}

\usepackage{xcolor}
\usepackage{algorithm}

\newcommand{\TCBlue}[1]{\textcolor{blue}{\ensuremath{\triangleright}\,#1}}

\newlength{\AlgShift}
\def\BibTeX{{\rm B\kern-.05em{\sc i\kern-.025em b}\kern-.08em
    T\kern-.1667em\lower.7ex\hbox{E}\kern-.125emX}}
\usepackage{eso-pic}

\AddToShipoutPictureBG*{
  \AtPageLowerLeft{%
    \raisebox{25pt}{\makebox[\paperwidth]{\begin{minipage}{21cm}\centering
    \fontsize{10}{8}\selectfont
    \end{minipage}}}%
  }
}

\definecolor{lightgray}{gray}{0.9}

\def\BibTeX{{\rm B\kern-.05em{\sc i\kern-.025em b}\kern-.08em
    T\kern-.1667em\lower.7ex\hbox{E}\kern-.125emX}}
\title{\LARGE \bf \vspace{6mm}
At the Limit: Autonomous Vehicle Drifting under Generative Policies}

\author{Jie He$^{1*\dagger}$, Meixun Qu$^{1*}$,
Cheng Hu$^{2}$, Lei Xie$^{2}$,
Ezio Bartocci$^{1}$, and Radu Grosu$^{1}$
\thanks{$^*$ \textbf{Equal contribution}.}
\thanks{$^1$ Technische Universität Wien, Vienna, Austria.}
\thanks{$^2$ Zhejiang University, Hangzhou, China.}
\thanks{$^\dagger$ Contact: {\tt\small jie.he@tuwien.ac.at}.}
}

\begin{document}

\newacronym{lidar}{LiDAR}{Light Detection and Ranging}
\newacronym{radar}{RADAR}{Radio Detection and Ranging}
\newacronym{mpc}{MPC}{Model Predictive Control}
\newacronym{mpcc}{MPCC}{Model Predictive Contouring Control}
\newacronym{iot}{IoT}{Internet of Things}
\newacronym{bev}{BEV}{Bird's-Eye View}

\newacronym{gpu}{GPU}{Graphics Processing Unit}
\newacronym{fps}{FPS}{Frames Per Second}
\newacronym{kf}{KF}{Kalman Filter}
\newacronym{ads}{ADS}{Autonomous Driving Systems}
\newacronym{iac}{IAC}{Indy Autonomous Challenge}
\newacronym{fsd}{FSD}{Formula Student Driverless}
\newacronym{rrt*}{RRT*}{Rapidly exploring Random Tree Star}
\newacronym{mgbt}{MGBT}{Multilayer Graph-Based Trajectory}
\newacronym{ftg}{FTG}{Follow The Gap}
\newacronym{gbo}{GBO}{Graph-Based Overtake}
\newacronym{rbf}{RBF}{Radial Basis Function}
\newacronym{map}{MAP}{Model- and Acceleration-based Pursuit}
\newacronym{sqp}{SQP}{Sequential Quadratic Programming}
\newacronym{roc}{RoC}{Region of Collision}
\newacronym{cpu}{CPU}{Central Processing Unit}
\newacronym{cots}{CotS}{Commercial off-the-Shelf}
\newacronym{obc}{OBC}{On-Board Computer}

\newacronym{sota}{SOTA}{State-of-the-Art}
\newacronym{qp}{QP}{Quadratic Programming}
\newacronym{A2RL}{A2RL}{Abu Dhabi Racing League}
\newacronym{GP}{GP}{Gaussian Process}
\newacronym{FITC}{FITC}{Fully Independent Training Conditional}
\newacronym{VFE}{VFE}{Variational Free Energy}
\newacronym{LQR}{LQR}{Linear Quadratic Regulator}
\newacronym{FSDP}{FSDP}{\textit{Fast and Safe Data-Driven Planner}}
\newacronym{gp}{GP}{Gaussian Process}
\newacronym{sgp}{SGP}{Sparse Gaussian Process}
\newacronym{VESC}{VESC}{Vedder Electronic Speed Controller}

\maketitle
\thispagestyle{fancy}


\pagestyle{empty}

\setlength{\tabcolsep}{3pt}
\glsresetall
\setlength{\abovedisplayskip}{6pt plus 1pt minus 1pt}
\setlength{\belowdisplayskip}{6pt plus 1pt minus 1pt}
\setlength{\abovedisplayshortskip}{3pt plus 1pt}
\setlength{\belowdisplayshortskip}{3pt plus 1pt}

%

\begin{abstract}
Autonomous drifting requires agile control despite tire-force saturation and highly nonlinear vehicle dynamics. We propose an online reinforcement learning method for flow policies that combines stochastic differential equation (SDE) sampling for exploration with value-based policy optimization. This method supports both training from scratch and fine-tuning a policy pretrained on expert demonstrations. After training, control actions can be generated using either
SDEs or ordinary differential equations (ODEs). Experiments on circular and
eight-shaped drifting in an F1TENTH digital twin platform show that both training strategies can achieve effective real-time drift control,
with tighter tracking, higher success rates, and better robustness under noisy conditions.
\end{abstract}

\section{Introduction}
Autonomous driving systems must remain controllable in safety-critical scenarios such as emergency obstacle avoidance or driving on low-friction roads, where the vehicle often operates at its physical limit and requires a maneuver to recover from the edge of instability. Such behavior lies at the core of \textit{drifting control}, in which the tires are deliberately driven into force saturation while the front wheels counter-steer to maintain a large sideslip angle. Although skilled human drivers can perform drifting reliably, designing an automatic controller that is robust and runs in real time on embedded hardware remains challenging.

Existing approaches in autonomous drifting control can be divided into model-based methods and learning-based methods. Model-based controllers~\cite{weber2023modeling, jonathan2024beyond}, such as MPC, formulate vehicle state and path tracking as an online optimal control problem. They provide theoretical guarantees and handle constraints explicitly, but require accurate vehicle models and are sensitive to uncertainty. The online optimization can be computationally expensive, and its real-time performance depends heavily on solver design, which depends on problem size and expert experience. 


Reinforcement learning (RL) enables end-to-end control, but often relies on heuristic environment and reward design and requires extensive interaction data. For drifting, precise coordination of drive torque and steering makes exploration particularly challenging: ineffective actions can cause early loss of control, limiting useful learning experience.

Expert demonstrations can provide an effective initialization for reinforcement learning, reducing ineffective exploration and facilitating subsequent improvement through online interaction. However, human drifting demonstrations are scarce and expensive to obtain. Moreover, the benefits of pretraining can vary across tasks. We therefore seek a training method that can distill expert demonstrations when available while retaining the ability to learn from scratch.

Generative modeling offers a flexible approach to policy design
by transforming a base distribution into a learned action
distribution through differential-equation-based
sampling approaches~\cite{chi2025diffusion,lipman2022flow}.
Its ability to represent multimodal action distributions
may support more effective exploration in complex driving
tasks.

However, training high-frequency generative policies with RL remains challenging, particularly because iterative action generation must be effectively integrated with online environment interaction. To address this challenge, we propose an online RL method that combines SDE-based action sampling with actor--critic policy optimization and adaptive control of the sampling noise. We further introduce a tractable path-wise entropy surrogate, computed from the conditional transition densities of the discretized SDE, to encourage exploration during online training without evaluating the generally intractable marginal density of the generated action. The proposed method supports both training from scratch and fine-tuning from an expert-pretrained policy.


We evaluate our approach on an F1TENTH digital-twin platform~\cite{o2020f1tenth,gordiienko2024experimental}. Both training strategies yield policies capable of stable circular and figure-eight drifting. Although trained solely on the nominal model without domain randomization, the resulting online policies maintain high success rates in \textit{zero-shot} evaluations under initial-state perturbations, vehicle-parameter variations, and dynamics noise, generally outperforming both the expert model-based controller and the policy obtained through imitation learning alone. With only four integration steps, the fine-tuned flow policy achieves an average inference latency of 0.5 ms on an Intel NUC13ANK, providing a speedup of more than $3\times$ over the model-based expert controller.

In summary, our main contributions are:
\begin{itemize}
    \item We present a novel generative RL framework for autonomous drifting, providing a unified pipeline that supports both training from scratch and offline-to-online fine-tuning from expert demonstrations.
    
    \item We design an online RL method for flow policies that combines SDE-based exploration, actor--critic optimization, and adaptive sampling noise.
          
    \item We systematically validate our method on an F1TENTH digital twin. Its simplicity, effectiveness, and robustness offer new insights into end-to-end control under extreme driving conditions.
\end{itemize}

\section{Related Work}
\noindent \textbf{Model-based Drift Control.} Prior work used error dynamics and model inversion to regulate drifting control inputs~\cite{hindiyeh2013dynamics,  goh2016simultaneous,goh2020toward}. MPC-based optimal control approaches have since become mainstream. \cite{hu2022combined} designed a linearized MPC controller for drifting control of a scale car. \cite{weber2023modeling, jonathan2024beyond} designed a more sophisticated nonlinear MPC controller for dynamic path and state tracking on a real car. However, MPC can require substantial computation to solve online optimization problems. Although tools such as CasADi~\cite{andersson2019casadi} are widely used, large problems often require customized solvers for real-time embedded deployment. The HDDC controller~\cite{chen2023dynamic} combines optimal control, error dynamics and model inversion through a modular design approach. In this work, we select it mainly due to its clear physical intuition. In principle, other model-based controllers can also provide expert demonstrations.

\noindent \textbf{Learning-based Drift Control.} \cite{cutler2016autonomous} presents an RL framework for drifting control. \cite{cai2020high} proposed a model-free approach using the Soft Actor-Critic (SAC) algorithm~\cite{haarnoja2018soft} for high-speed drift control. \cite{zhao2023adaptive} developed the JS-SAC algorithm to incorporate expensive human drifting data for better learning efficiency. Recently, \cite{djeumou2025reference, zhou2025learning} leveraged the parallel capability of the PPO algorithm~\cite{schulman2017proximal} to train an RL-based drifting controller from scratch. However, this line of work still relies on large amounts of training data. Moreover, these works directly apply existing RL algorithms, while we propose a new RL algorithm that supports both online learning from scratch and offline-to-online transfer.

\noindent \textbf{Generative Policies.}
Early imitation learning methods employ behavior
cloning~\cite{torabi2018behavioral} or adversarial
learning~\cite{ho2016generative}, but can suffer from
compounding errors or training instability, respectively.
Inspired by diffusion models for image
generation~\cite{ho2020denoising}, diffusion policies
provide an effective framework for learning complex
behaviors from demonstrations~\cite{chi2025diffusion}.
Flow policies further employ flow
matching~\cite{lipman2022flow} to enable action generation
with fewer sampling steps~\cite{park2025flow}.

Beyond imitation learning, recent studies explore offline-to-online RL
for flow policies~\cite{park2025flow,wagenmaker2025steering,li2026reinforcement}.
These methods use ODE-based action generation with limited stochasticity. In contrast, our
method introduces SDE-based sampling during online training
and supports both fine-tuning a pretrained policy and learning from scratch. Diffusion models have also been used to
generate diverse vehicle parameters~\cite{djeumou2024one},
which addresses vehicle modeling rather than policy learning.

\vspace{-0.8em}
\section{Vehicle Dynamics and Control}
\label{sec:pre}

\subsection{Vehicle Modeling}
\vspace*{-0.5ex}
\begin{figure}[h]
    \centering
    \vspace*{-3ex}
    \includegraphics[width=0.4\textwidth]{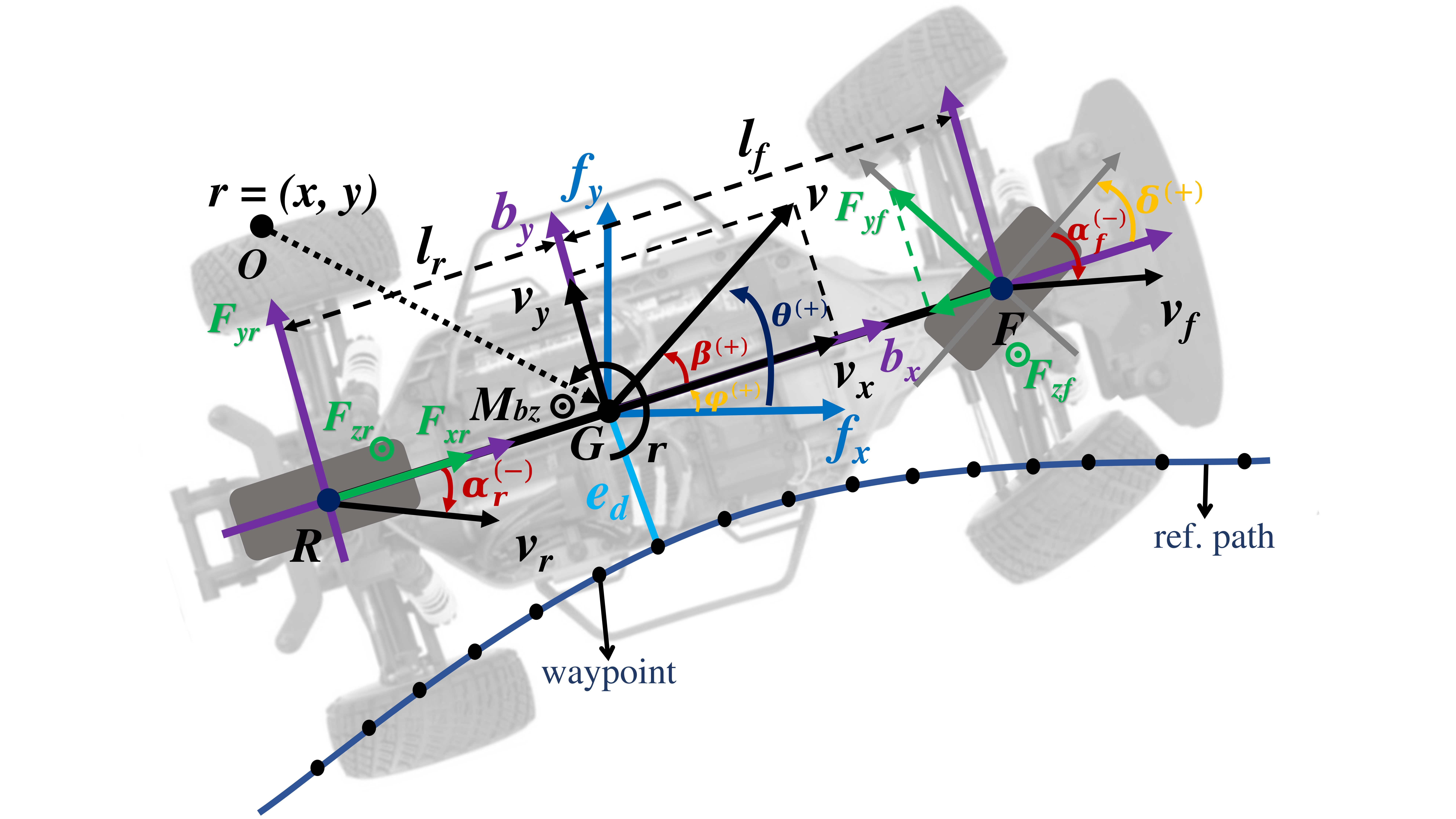}
    \vspace*{-2ex}
    \caption{Dynamics Model of an RWD F1TENTH Car}
    \vspace*{-2ex}
    \label{fig:dynamics}
\end{figure}

\subsubsection{Vehicle Model}
As shown in Fig.~\ref{fig:dynamics}, this work employs a 3-DoF single-track bicycle model:
\begin{gather}
\label{equ:vehicle_dynamics}
\begin{array}{l}
\dot \beta = \frac{1}{mv}[F_{yf}\text{cos}(\delta-\beta)+F_{yr}\text{cos}\beta-F_{xr}\text{sin}\beta]-r  \\[7pt]
\dot r=\frac{1}{I_{z}}(l_{f}F_{yf}\text{cos}\delta-l_{r}F_{yr})  \\[7pt]
\dot v=\frac{1}{m}[F_{xr}\text{cos}\beta-F_{yf}\text{sin}(\delta-\beta)+F_{yr}\text{sin}\beta]  \\[7pt]
\end{array}
\end{gather}
where $\beta$, $r$ and $v$ are the sideslip angle, yaw rate, and total velocity at the center of gravity. Vehicle motion is governed by the front lateral force $\bm{F}_{yf}$ and the rear lateral and longitudinal forces $\bm{F}_{yr}$ and $\bm{F}_{xr}$. The lateral tire forces are determined by the tire slip angles $\alpha_f$/$\alpha_r$.


\subsubsection{Tire Model}
During drifting, lateral tire forces typically saturate as the tire slip angle increases. To capture this nonlinear behavior and to model the coupling of longitudinal and lateral forces in the rear wheel, we adopt the Fiala brush tire model \cite{goh2020toward, hindiyeh2009equilibrium, hindiyeh2013dynamics} as shown below:
\begingroup
\setlength{\arraycolsep}{0.5pt}
\begin{gather}
F_{yi}= \left\{\begin{array}{ll} 
-C_{\alpha i} z_{i} + \frac{C_{\alpha i}^{2}}{3\xi_i\mu F_{zi}} \left| z_{i} \right| z_{i} - \frac{C_{\alpha i}^{3}}{27 \xi_{i}^{2} \mu^2 F_{zi}^{2}}z^{3}_{i},  & \left| \alpha_{i} \right|\le \alpha_{sli} \\ 
-\xi_i \mu F_{zi} \operatorname{sgn} (\alpha_{i}), & \left| \alpha_{i} \right| > \alpha_{sli}
\end{array} \right.
\label{equ:Fiala}
\end{gather}
\endgroup
where $z_i=\text{tan}\alpha_i$, $i=f$ for $F_{yf}$ and $i=r$ for $F_{yr}$. $C_{\alpha_i}$ is the tire cornering stiffness. $F_{zi}$ is the normal force of the tire, and $\mu$ is the friction coefficient. For the front wheel $\xi_f=1$, and for the rear wheel $\xi_r=\frac{\sqrt{(\mu F_{zr})^2-F^2_{xr}}}{\mu F_{zr}}$.

The Fiala model computes lateral tire forces from $\alpha_i$, with saturation slip angle $\alpha_{sli}=\text{arctan}\frac{3\mu\xi_i F_{zi}}{C_{\alpha_{i}}}$. If $|\alpha_i|\le \alpha_{sli}$, the tire is unsaturated, and we use the first formula in Eq.~\ref{equ:Fiala} to compute lateral force; otherwise, the tire saturates and slides.




\subsubsection{Reference Design via Equilibrium}
We first design reference trajectories, and uniformly distribute reference points along them. At each waypoint, the reference position $x_r$, $y_r$ and the ideal course angle $\theta_r$ are known. We also assign $\beta_r$ and velocity $v_r$ to each point.

For steady-state drifting, $\dot \beta=\dot r=\dot v=0$. Equation~\ref{equ:vehicle_dynamics} then gives three equations with five unknowns $(\beta^{\mathrm{eq}},r^{\mathrm{eq}},v^{\mathrm{eq}}, \delta^{\mathrm{eq}}, F_{xr}^{\mathrm{eq}})$. Given $\beta_r$ and $v_r$ at each reference point, we can solve for the remaining three variables $(r^{\mathrm{eq}},\delta^{\mathrm{eq}}, F_{xr}^{\mathrm{eq}})$, and set ${r_r}=r^{\mathrm{eq}}$. 

We further determine the yaw angle as $\varphi=\theta-\beta$. With $v_x=v\text{cos}\beta$, $v_y=v\text{sin}\beta$, we can obtain the \textit{reference state} ($\beta_r, r_r, v_{xr}, v_{yr}, v_r$) and the \textit{reference pose} ($x_r, y_r, \varphi_r, \theta_r$).
From a dynamics perspective, the equilibrium lies near an unstable (saddle-type) regime, where small perturbations can be amplified, making steady drift intrinsically challenging.
\subsection{Collection of Expert Demonstrations}
\subsubsection{Simulation Procedure}
At each step, we simulate the vehicle dynamics model in Eq.~\ref{equ:vehicle_dynamics} with the \textit{Euler method}, updating the state ($\beta, r, v_x, v_y, v$) and pose ($x, y,\varphi, \theta$). Next, we search forward for the nearest reference point and extract the reference pose/state. We then compute the errors between the current and reference states, as well as between the current and reference poses, which are used to compute the rewards.
%
\subsubsection{Experience Collection}
We assemble the observation vector as 
$
    \bm{o_t} = (x, y, \varphi, s, \beta,r, v_x, v_y, e_{\beta}, e_r, e_d, e_\theta)
$ at each simulation step,
where $s$ is the distance traveled so far, $e_\beta, e_r, e_\theta$ are the errors between the vehicle's current and reference states, defined as $e_p=|p-p_r|$ for $p\in\{\beta,r,\theta\}$. $e_d$ is the lateral error that characterizes the distance between the vehicle's current position and the reference position, and it is given by $e_d=[x-x_r, y-y_r] \cdot [-\text{sin}\theta, \text{cos}\theta]^T$.

We combine the current observation and $L$ previous observations into $\bm{s_t} = (\bm{o_t}, \bm{o_{t-1}}, ..., \bm{o_{t-L}})$ where $L=9$ and $\bm{s_t}$ is the observation state at time $t$. We then input $\bm{o_t}$ to the HDDC controller and obtain the action as $\bm{a_t}=(\delta_t, F_{xr,t})$. Simulating the next step gives $\bm{o_{t+1}}$, $\bm{s_{t+1}}$, and reward $r_t$ (Section~\ref{sub:reward_design}). Repeating this procedure with HDDC yields an expert dataset $\mathcal{D}$ of transitions $(\bm{s_t},\bm{a_t},r_t,\bm{s_{t+1}})$.

To diversify experience from the deterministic model in Eq.~\ref{equ:vehicle_dynamics}, we add noise to $\beta$, $r$, and $v$ during integration.

\section{Flow Policy}
\label{sec:flow_policy}

In this paper, we consider an infinite-horizon, partially observable Markov decision process (POMDP)
\((\mathcal{S}, \mathcal{A}, \omega, \mathcal{T}, r, \gamma)\),
where \(\mathcal{S}\) is the state space, \(\mathcal{A}\) is the action space,
\(\mathcal{T}:\mathcal{S}\times\mathcal{A}\to\Delta(\mathcal{S})\) is the transition kernel
(with \(\Delta(\cdot)\) denoting the set of probability measures over a set),
\(r:\mathcal{S}\times\mathcal{A}\to\mathbb{R}\) is the reward function,
\(\omega\in\Delta(\mathcal{S})\) is the initial-state distribution,
and \(\gamma\in[0,1)\) is the discount factor.
We also assume access to a prior offline dataset
\(\mathcal{D}=\{(s_t,a_t,r_t,s_{t+1})\}\) consisting of transition rollouts collected from the HDDC controller.
The goal of offline-to-online RL is to find a policy
\(\pi(a|s)\in\Delta(\mathcal{A})\) that maximizes the expected discounted return: $R(\pi)\triangleq
\mathbb{E}_{\substack{
s_0\sim \omega,\;
a_t\sim \pi(\cdot\mid s_t),\\
s_{t+1}\sim \mathcal{T}(\cdot\mid s_t,a_t)
}}
\Bigg[\sum_{t=0}^{\infty}\gamma^{t}\, r(s_t,a_t)\Bigg].$

Inspired by the \textit{Flow Matching} approach in generative modeling, we use a \textit{Flow model} or a \textit{Diffusion model} to generate actions, rather than adopting a simple Gaussian policy commonly used in RL.
\subsection{Flow Matching}
\subsubsection{Models}

A \textit{Flow Model} is an ODE $dX_t=u_t(X_t)dt$, with $X_0\sim p_{init}$ and $X_t\sim p_t$ for $t\in[0,1]$. The vector field $u_t:[0,1]\times\mathbb{R}^d\to\mathbb{R}^d$ transports the initial distribution (e.g., a unit Gaussian) along the \textit{marginal probability path} $p_t$.

The corresponding \textit{Diffusion Model} is the SDE $dX_t=[u_t(X_t)+\frac{\sigma_t^2}{2}\nabla\log p_t(X_t)]dt+\sigma_t dW_t$, where $W_t$ is Brownian motion. The \textit{marginal score function} $\nabla\log p_t$ compensates for the diffusion term, preserving the same probability path $p_t$ for different diffusion coefficients $\sigma_t$.

\subsubsection{Action Generation}
\label{subsub:action_generation}
We use neural networks $\theta$ to learn the vector field $u^\theta_t(X_t)$ and score function $s^\theta_t(X_t)$. 
The corresponding ODE/SDE become $dX_t=u^\theta_t(X_t)dt$ and $dX_t=[u^\theta_t(X_t)+\frac{\sigma^2_t}{2}s^\theta_t(X_t)]dt + \sigma_t dW_t$.

For the drifting task, $X_t$ represents the steering angle $\delta$ and longitudinal force $F_{xr}$. Given updated car states, we initialize $X_0\sim\mathcal{N}(0,I_d)$ with $d=2$ and integrate the learned ODE/SDE to obtain action $X_1$. 
\subsubsection{Training from Data} The original goal of flow matching is to make $X_1$ follow $p_{\text{data}}$, where $p_{\text{data}}$ characterizes the distribution of ``expert actions'' in our case. To this end, given a flow model $dX_t=u^\theta_t(X_t)dt$, we can regress $u^\theta_t(X_t)$ against $u_t^{\text{target}}(X_t|z)$ with the following loss function:
$
\mathcal{L}_{CFM}(\theta) = \mathbb{E}
[\left\| \, u^{\theta}_t(X_t) - 
u_t^{\text{target}}(X_t|z)
\right\|_{2}^{2}]
$.


To get the closed form of $u_t^{\text{target}}\bigl(X_t|z\bigr)$, it is common to set the conditional probability path as \textit{Gaussian} such that $X_t \sim p_t(\cdot|z)=\mathcal{N}(\alpha_tz,\beta_t^2 I_d)$. In this paper, we set $\alpha_t=t$ and $\beta_t=1-t$. This gives the standard Gaussian $p_0(\cdot|z)=\mathcal{N}(0,I_d)$ and the Dirac distribution $p_1(\cdot|z)=\delta_z$. Consequently, the regression loss becomes:
$
\small
\mathcal{L}_{CFM}(\theta) = \mathbb{E}_{\square}
\Bigl[
  \bigl\|
    u^{\theta}_t\bigl(\alpha_t z + \beta_t \epsilon\bigr)
    - \bigl(\dot{\alpha}_t z + \dot{\beta}_t \epsilon\bigr)
  \bigr\|^{2}_2,
\Bigr]
$
where $\square = \{\,t \sim \mathrm{Unif}([0,1]),\; z \sim p_{\text{data}},\; \epsilon \sim \mathcal{N}(0, I_d)\,\}$.
In particular, when $\alpha_t = t$, $\beta_t = 1-t$, $\dot{\alpha}_t = 1$, and $\dot{\beta}_t = -1$:
\begin{equation}
\label{equ:cfm_final}%
\mathcal{L}_{CFM}(\theta) = \mathbb{E}_{\square}
\bigl[\left\| \, u^{\theta}_t\bigl(t z + (1-t)\epsilon\bigr) -
\bigl(z-\epsilon\bigr)\right\|^{2}_2\bigr].
\end{equation}

Once $u^\theta_t(x_t)$ is well trained, we can use the parametrized SDE to generate actions, without the need to train a separate score function. The conversion formula directly computes $s^\theta_t(x_t)$ from $u^\theta_t(x_t)$:
$s^\theta_t(x_t)=\frac{\alpha_tu^\theta_t(x_t)-\dot{\alpha_t}x_t}{\beta^2_t\dot{\alpha_t}-\alpha_t\dot{\beta_t}\beta_t}
\label{equ:conversion_formula}~$\cite{holderrieth2025introduction}.

\subsection{Offline Imitation Learning}
\begin{algorithm}[t]
\caption{Flow ODE}
\footnotesize
\begin{algorithmic}[1]
\Function{$FlowODE$}{$s$}  \Comment{$\pi_{\theta}^{\text{ODE}}$}
\State $x_0\sim\mathcal{N}(0, I_d)$, $h=1/M$
  \For{$i = 0,1,..., M-1$}
    \State $t=i/M$
    \State $x_{t+h} \gets x_{t} + h \cdot u_\theta (x_{t}, t, s)$ 
  \EndFor
  \State \Return $x_1$
\EndFunction
\label{alg:flow_ode}
\end{algorithmic}
\end{algorithm}
\begin{algorithm}[t]
\caption{Offline Training}
\footnotesize
\begin{algorithmic}[1]
\State \textbf{Input:} Offline dataset $\mathcal{D}=\{(s,a,r,s')\}$.
\While{not converged}
  \State Sample batch $\mathcal{B}=\{(s, a, r,s')\} \subseteq \mathcal{D}$.
  \State \TCBlue{Train critic $Q_\phi$ with $\mathcal{B}$}.
  \State Sample $(s,a,s')\sim \mathcal{B}, a'\sim \pi_\theta^{\text{ODE}}(s')$.
  \State For each critic $Q_{\phi_k}$, compute Eq.~\ref{equ:pretrain_critic_loss}.
  \State Update $\phi$ by minimizing $L(\phi)=\sum_{k}^{}L(\phi_k)$.
  \State Update target networks with $\overline{\phi}_k \gets \tau \phi_k + (1-\tau)\overline{\phi}_k$.
  
  \State \TCBlue{Train actor $\pi_\theta^{\text{ODE}}$ with $(s,a)$ from $\mathcal{B}$}.
  \State Sample $s \sim \mathcal{B}$.
  \State Compute \textit{Q loss} via $L_q(\theta)=-\mathbb{E}_{s \sim \mathcal{B}}
    \frac{1}{K} \cdot\sum\limits_{k=1}^{K} Q_{\phi_k}(s, \pi_\theta^{\text{ODE}}(s))$
  
  \State Sample $(s,a) \sim \mathcal{B}$, assign $z \gets a$.
  \State Sample $t \sim \text{Unif}([0,1]), \epsilon \sim \mathcal{N}(0, I_d)$.
  \State Compute \textit{BC loss} via $L_{\text{BC}}(\theta) = \mathbb{E}_{\square_1}\left\| u_\theta(x_t, t,s) - (z-\epsilon) \right\|_2^2.$
  \State Update $\theta$ by minimizing $L(\theta)=L_q(\theta)+\beta L_{\text{BC}}(\theta)$.
\EndWhile
\State \textbf{Output:} Pretrained weights $\theta_1$, $\phi_1$ and $\overline{\phi}_1$.
\label{alg:pretrain}
\end{algorithmic}
\end{algorithm}
\setlength{\textfloatsep}{0.1cm}
\setlength{\floatsep}{0.1cm}

\subsubsection{Network Design}
We adopt an actor-critic training paradigm. For the actor, we define the policy as $\pi_{\theta}^{\text{ODE}}(s)\triangleq \text{FlowODE}(s)$. As shown in Alg.~1, the output action is initialized with $x_0\sim\mathcal{N}(0, I_d)$. In Lines 3--5, the \textit{Euler method} integrates the intermediate action for $M$ steps of size $h$, and $u_\theta (x_{t}, t, s): \mathbb{R}^\mathcal{|A|} \times [0, 1]  \times \mathcal{S} \to \mathbb{R}^\mathcal{|A|} (\mathcal{|A|}=2)$ is the vector field we aim to learn. Given the state, intermediate action, and time, it outputs the velocity field. The algorithm returns $x_1$ as the generated action.

We also train critic networks $Q_{\phi_k}(s,a): \mathcal{S} \times \mathcal{A}\to\mathbb{R}$, to learn state-action values from expert traces. The ensemble of $K$ critics has parameters $\phi=(\phi_1, \ldots, \phi_K)$. 

In this work, we use three-layer multilayer perceptrons (MLPs) with 256 hidden units to construct $u_\theta$ and $Q_{\phi_k}$.
 
\subsubsection{Offline Algorithm}
Alg.~2 trains the actor and critics on batches $\mathcal{B}=\{(s,a,r,s')\}$ of expert experience. Each critic minimizes:
\vspace{-1em}
\begin{equation}
\footnotesize
L(\phi_k)=\mathbb{E}_{\substack{(s,a,s')\sim \mathcal{B}\\ a'\sim \pi_\theta^{\text{ODE}}(s')}}
\Big[\,Q_{\phi_k}(s,a)-\Big(r + \gamma\cdot \tfrac{1}{K}\sum\limits_{k=1}^{K} Q_{\overline{\phi}_k}(s', a')\Big)\Big]^2
\label{equ:pretrain_critic_loss}
\end{equation}
where $Q_{\overline{\phi}_k}$ are target critic networks, and as shown in Line 8 in Alg.~2, we apply a soft update of their parameters. 

We learn the actor with two loss functions. The first is the \textit{Q loss} that aims to maximize the state-action value: $
    L_q(\theta)=-\mathbb{E}_{s \sim \mathcal{B}}
    \frac{1}{K} \cdot\sum\limits_{k=1}^{K} Q_{\phi_k}(s, \pi_\theta^{\text{ODE}}(s))$. The second is the \textit{BC loss} $L_{\text{BC}}(\theta)$ that follows Eq.~\ref{equ:cfm_final}. We sample $(s,a) \sim \mathcal{B}$, set $z=a$, and sample $t \sim \text{Unif}([0,1]), \epsilon \sim \mathcal{N}(0, I_d)$ to construct one sample of $x_t$ as $x_t \gets tz + (1-t)\epsilon$. With Eq.~\ref{equ:cfm_final}, $L_{\text{BC}}(\theta)$ can be written as: $L_{\text{BC}}(\theta) = \mathbb{E}_{\square_1}\left\| u_\theta(x_t, t,s) - (z-\epsilon) \right\|_2^2$, where $\square_1$ is the sampling process in Lines 12--13 of Alg.~2. Combining $L_q(\theta)$ and $L_{\text{BC}}(\theta)$ yields the complete actor loss: $L(\theta)=L_q(\theta)+\beta L_{\text{BC}}(\theta)$, where $\beta$ determines the importance of behavior cloning in $L(\theta)$.

\subsection{Online Learning}
\subsubsection{ODE-SDE Transition}
The flow policy might still be suboptimal after offline training as it lacked interaction with the environment. Moreover, the action still has limited diversity because the policy was generated via an ODE. To mitigate this issue, we switch from ODE to SDE action generation for richer exploration.


To maximize reward while controlling SDE noise, we formulate online training as a path-entropy-regularized RL problem: $\max_{\pi}\ \mathbb{E}_{\pi}\!\left[\sum_{t} r(s_t,a_t)\right]$ with constraint $\mathbb{E}_{(s_t,a_t)\sim \rho_{\pi}}\!\big[-\log(\pi_t(a_t|s_t))\big] \ge \mathcal{H}_0$, where $\mathcal{H}_0$ is the desired entropy.

We define the policy in online training as $\pi_{\theta}^{\text{SDE}}(s)\triangleq \text{FlowSDE}(s)$. Alg.~3 describes how we generate actions with an SDE and compute the associated entropy surrogate.

In Line 5, we compute the score function $s_\theta (x_{t}, t, s): \mathbb{R}^\mathcal{|A|} \times [0, 1]  \times \mathcal{S} \to \mathbb{R}^\mathcal{|A|}$, from the learned vector field by using the conversion formula. 

In Line 6, the noise schedule is $\sigma_t=\sqrt{\alpha \frac{1-t+h}{t+h}}$~\cite{domingo2024adjoint}. The schedule decreases with $t$, encouraging exploration early in action generation and reducing randomness thereafter. The learnable parameter $\alpha$ controls its scale. By combining the learned vector field $u_\theta$, score function $s_\theta$ and $\sigma_t$, we get the new vector field of the SDE as $ v_\theta (x_{t}, t, s): \mathbb{R}^\mathcal{|A|} \times [0, 1]  \times \mathcal{S} \to \mathbb{R}^\mathcal{|A|}$ (Line 7).

In Line 8, we use the \textit{Euler--Maruyama method} to simulate the SDE. As shown in Line 9, at each integration step, $x_{t+h}$ actually follows a normal distribution with mean $\overline{x_{ct}}=x_t + h \cdot v_\theta(x_t, t,s)$, and standard deviation $\sigma_{ct}=\sqrt{h}\cdot \sigma_t$. We compute a Monte Carlo estimate of the path-wise entropy surrogate in a single integration step as: $\mathcal{H}_t = -\text{log}\mathcal{N}(x_{t+h}|\overline{{x}_{ct}},\sigma_{ct}^2)=\frac{(x-\overline{{x}_{ct}})^2}{2\sigma_{ct}^2}+\text{log}(\sqrt{2\pi}\sigma_{ct})
\label{equ:entropy_Ht}$. We sum the entropy contributions in Line 11 and normalize the result by the number of integration steps $M$ and action dimension $d$ in Line 12. Finally, Alg.~3 returns the output action $x_1$ and the normalized entropy surrogate $\mathcal{H}$. Note that $\mathcal{H}$ can be determined by $\alpha$, so with the actor $\pi_{\theta}^{\text{SDE}}$, we can design a maximum-entropy RL algorithm for online learning.
\begin{algorithm}[t]
\caption{Flow SDE}
\footnotesize
\begin{algorithmic}[1]
\Function{$FlowSDE$}{$s$} \Comment{$\pi_{\theta}^{\text{SDE}}$}
\State $x_0\sim\mathcal{N}(0, I_d)$, $h=1/M$, $\mathcal{H} \gets 0$
  \For{$i = 0,1,..., M-1$}
    \State $t=i/M$
    \State $s_\theta(x_t, t, s) = \frac{\alpha_t u_\theta(x_t,t,s)- \dot{\alpha_t}x_t}{\beta^2_t\dot{\alpha_t}-\alpha_t\dot{\beta_t}\beta_t} =
    \frac{tu_\theta(x_t,t,s)-x_t}{1-t}$ 
    \State $\sigma_t=\sqrt{\alpha \cdot\frac{1-t+h}{t+h}}, \alpha>0$
    \State $ v_\theta(x_t,t,s) = u_\theta(x_t,t,s)+\frac{1}{2}\sigma_t^{2}s_\theta(x_t,t,s)$
    \State $x_{t+h}\gets \underbrace{x_t + h \cdot v_\theta(x_t, t,s)}_{\color{blue}{\overline{x_{ct}}}}+ \underbrace{\sqrt{h}\cdot \sigma_t }_{\color{blue}{\sigma_{ct}}}\epsilon_t, \epsilon_t\sim\mathcal{N}(0, I_d)$
    \State $x_{t+h} \sim \mathcal{N}(x_{t+h}|\overline{{x}_{ct}}, \sigma_{ct}^2)$
    \State Compute $\mathcal{H}_t \gets -\text{log}\mathcal{N}(x_{t+h}|\overline{{x}_{ct}},\sigma_{ct}^2)$ via Eq.~\ref{equ:entropy_Ht}.
    \State $\mathcal{H} \gets \mathcal{H} + \mathcal{H}_t$
    
  \EndFor
  \State $\mathcal{H} \gets \mathcal{H}/(M \cdot d)$
  \State \Return $x_1, \mathcal{H}$
\EndFunction
\label{alg:flow_sde}
\end{algorithmic}
\end{algorithm}

\begin{algorithm}[t]
\caption{Online Training}
\footnotesize
\begin{algorithmic}[1]
\If{pretrained weights are available}
    \State Initialize the velocity field, critics, and target critics with
    $u_{\theta_1}$, $Q_{\phi_1}$, and $Q_{\bar{\phi}_1}$, respectively.
\Else
    \State Randomly initialize the velocity field and critics, and initialize
    the target critics from the critics.
\EndIf
\State Initialize empty replay buffer $\mathcal{R}$.
\State Warm-up phase: run environment for $E_c$ episodes with SDE policy $\pi_{\theta}^{\text{SDE}}(s)$, and add experience $(s, a, r, s')$ to $\mathcal{R}$.

\For {$e=1, 2, ..., E_o$}
\State Receive initial observation state $s_1$.
\For{$t = 1, 2, ..., T$}
  \State \TCBlue{Interact with environment}.
  \State Get action $a_t=\pi_{\theta}^{\text{SDE}}(s_t)$.
  \State Execute $a_t$, get reward $r_t$ and $s_{t+1}$ through state transition.
  \State Add $(s_t, a_t, r_t, s_{t+1})$ to replay buffer $\mathcal{R}$.
  
  \State Sample batched experience $\mathcal{B}=\{(s, a, r,s')\}$ from $\mathcal{R}$.
  \State \TCBlue{Train critic $Q_\phi$ with $\mathcal{B}$}.
  \State Sample $(s,a,s')\sim \mathcal{B}$, get $a', \mathcal{H'} \gets \pi_\theta^{\text{SDE}}(s')$.
  \State For each critic $Q_{\phi_k}$, compute Eq.~\ref{equ:finetune_critic_loss}.
  \State Update $\phi$ by minimizing $L(\phi)=\sum_{k}^{}L(\phi_k)$.
  \State Update target networks with $\overline{\phi}_k \gets \tau \phi_k + (1-\tau)\overline{\phi}_k$.
  
  \State \TCBlue{Train actor $\pi_\theta^{\text{SDE}}$ with $(s,a)$ from $\mathcal{B}$}.
  \State Sample $s \sim \mathcal{B}$, get $a, \mathcal{H} \gets \pi_\theta^{\text{SDE}}(s)$.
  \State Update $\theta$ by minimizing Eq.~\ref{equ:finetune_q_loss}.
  \State Update $\alpha$ by minimizing Eq.~\ref{equ:alpha_loss}.
\EndFor
\EndFor
\State \textbf{Output:} Fine-tuned weights $\theta_2$, $\phi_2$ and $\overline{\phi}_2$.
\label{alg:finetune}
\end{algorithmic}
\end{algorithm}
\setlength{\textfloatsep}{0.1cm}
\setlength{\floatsep}{0.1cm}

\subsubsection{Online Algorithm}
Alg.~4 shows online training. We initialize the actor, critics, and target critics (Lines 1-4), create an empty replay buffer $\mathcal{R}$ (Line 5), and collect warm-up experience (Line 6).

We then train the actor and critics for $E_o$ episodes of $T$ steps. Each episode starts from a reset state $s_1$. At each step, we add a transition to $\mathcal{R}$ and train on a sampled batch $\mathcal{B}=\{(s,a,r,s')\}$.

The loss function for each critic is:
\begin{equation}
\small
L(\phi_k)=\mathbb{E}_{{\square}_2}
\Big[\,Q_{\phi_k}(s,a)-\Big(r + \gamma\cdot (\min_{k}Q_{\overline{\phi}_k}(s', a')+\mathcal{H}')\Big)\Big]^2
\label{equ:finetune_critic_loss}
\end{equation}
where $\square_2$ represents the sampling process in Line 16 of Alg.~4. To mitigate the overestimation of $Q$ values, here we only keep the minimum value from the target critics.

The actor loss maximizes both the $Q$ value and the entropy surrogate to encourage exploration:
\begin{equation}
\footnotesize
        L(\theta)=-\mathbb{E}_{\substack{s\sim \mathcal{B}\\ a, \mathcal{H} \gets \pi_\theta^{\text{SDE}}(s)}}
\Big[\min_{k}Q_{\phi_k}(s, a)+\mathcal{H}\Big]
\label{equ:finetune_q_loss}
\end{equation}

Since the learnable parameter $\alpha$ affects the entropy of action, we also add the following loss function for $\alpha$:
\begin{equation}
    L(\alpha) = \mathbb{E}[\alpha (\mathcal{H}-\mathcal{H}_0)]
\label{equ:alpha_loss}
\end{equation}
where $\mathcal{H}_0$ is the target entropy. As $\mathcal{H}_0$ is often small, when $L(\alpha) \to 0$, $\alpha \to 0$. From Alg.~3, this makes $\sigma_t \to 0$ (Line 6) and $v_\theta \to u_\theta$ (Line 7). In this case, $\pi_\theta^{\text{SDE}}$ reduces to $\pi_\theta^{\text{ODE}}$.

\section{Experiments}
\label{sec:experiment}
\setlength{\textfloatsep}{5pt plus 2pt minus 2pt}
\setlength{\floatsep}{5pt plus 2pt minus 1pt}
\setlength{\intextsep}{6pt plus 2pt minus 1pt}
\setlength{\dbltextfloatsep}{5pt plus 2pt minus 2pt}
\setlength{\dblfloatsep}{5pt plus 2pt minus 1pt}
\captionsetup{font=footnotesize,skip=3pt}

\subsection{Environment Setup}
\subsubsection{Reward Design}
\label{sub:reward_design}
We consider four types of rewards as follows: $r=r_{\rm pose}+r_{\rm state}+r_{\rm smooth}+r_{\rm term}$. With normalized errors $\bar e_m=e_m/e_{m,\max}$ and $f(z)=\exp(-|z|)-1$, pose and state tracking rewards use $r_{\rm pose}=w_{e_d}f(\bar e_d)+w_{e_\theta}f(\bar e_\theta)$ and $r_{\rm state}=w_{e_\beta}f(\bar e_\beta)+w_{e_r}f(\bar e_r)+w_{e_{v_x}}f(\bar e_{v_x})$. Smoothness is penalized by $r_{\rm smooth}=w_s(r_{\dot\beta}+r_{\dot r})+w_{\dot{u}}(r_{\dot\delta}+r_{\dot F_{xR}})+w_{\ddot{u}}(r_{\ddot\delta}+r_{\ddot F_{xR}})$, using the same exponential form and corresponding normalization scales. 

For the parameters, we simply set $w_{e_d}=w_{e_\theta}=w_{e_r}=1$, $w_{e_\beta}=w_{e_{v_x}}=2$, $w_s=0.2$, and $w_{\dot{u}}=w_{\ddot{u}}=0.1$. Early failure incurs $r_{\rm term}=-100$; otherwise, it is zero.

\subsubsection{Randomized Vehicle Model}
The parameters of the F1TENTH car can be found in~\cite{gordiienko2024experimental}. We set the simulation time step $dt=0.02s$ with a $50$ Hz control frequency. 

We consider three layers of disturbance: (1) \textit{Integration Layer.} We add the following noise to Eq.~\ref{equ:vehicle_dynamics} to account for unmodeled dynamics: $\bm{x}_{k+1} = \bm{x}_k + \bm{f}(\bm{x},\bm{u})dt + \bm{\Sigma}\sqrt{dt}\epsilon$, where $ \bm{\Sigma}=\text{diag}(\sigma_\beta, \sigma_r, \sigma_{v_x})$, and the standard deviation at each step is around $3\%-4\%$ of $\beta^{\mathrm{eq}}$, $r^{\mathrm{eq}}$ and $v_x^{\mathrm{eq}}$. (2) \textit{Force Layer.} We consider the unmodeled dynamics of the Fiala model by adding colored noise to the tire forces with the OU process: $dd_t=-\frac{1}{\tau}d_tdt+\sigma_cdW_t, d_t \sim \mathcal{N}(0, \frac{\tau\sigma_c^2}{2})$, where $d_t$ is the disturbance term at each time step. We set $\tau$ to $0.1s$, and $\sqrt{\tau\sigma_c^2/2}$ is around $1\%$ of the normal forces on the tires. (3) \textit{Parameter Layer.} We inject OU noise into $C_{\alpha f}$, $C_{\alpha r}$, and $\mu$. By setting $\tau=0.01s$, and $\sqrt{\tau\sigma_c^2/2}$ as $0.1\%$ of the $C_{\alpha f}$, and $C_{\alpha r}$ and $\mu$ values, we can characterize the reasonable parameter changes within the given simulation period.

\begin{figure*}[!t]
\centering
\includegraphics[width=.92\textwidth]{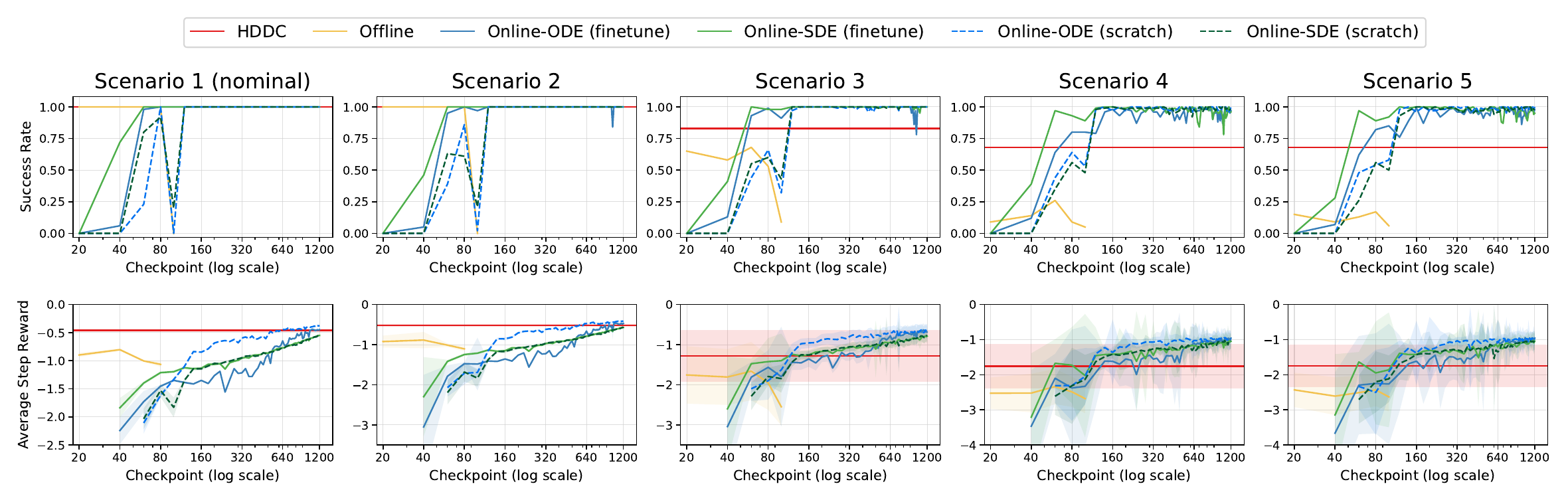}
\caption{Learning Curves for Circular Drift. Rewards are shown when success rate is larger than 0.}
\label{fig:circle_overall}
\vspace{1pt}
\includegraphics[width=.92\textwidth]{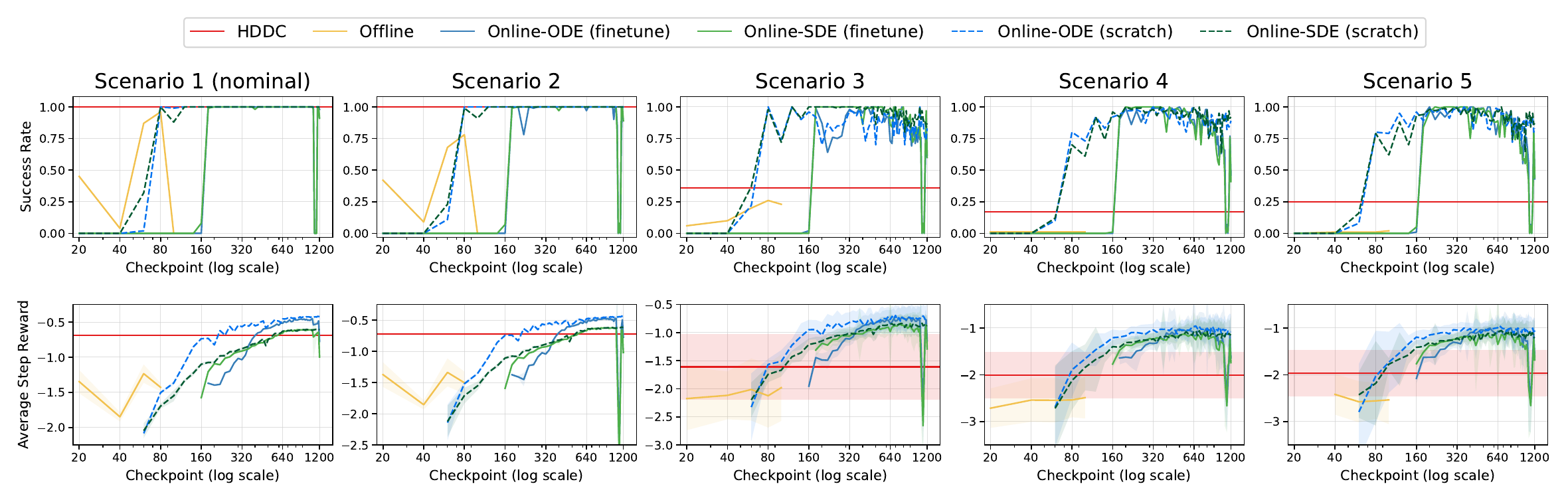}
\caption{Learning Curves for Eight-shaped Drift. Rewards are shown when success rate is larger than 0.}
\label{fig:eight_overall}
\end{figure*}

\begin{figure}[!t]
\centering
\includegraphics[width=.88\columnwidth]{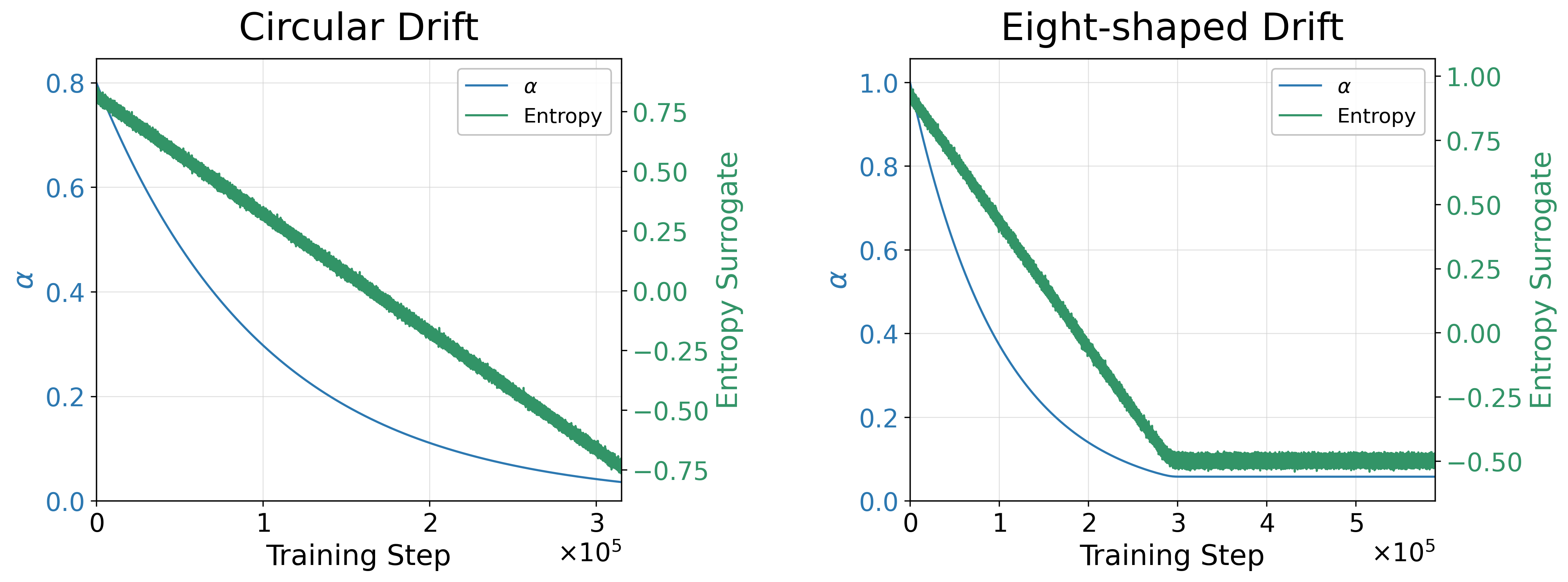}
\caption{Adaptive $\alpha$ and Entropy Surrogate}
\label{fig:alpha_entropy}
\end{figure}

\begin{figure*}[!t]
\centering
\includegraphics[width=.82\textwidth]{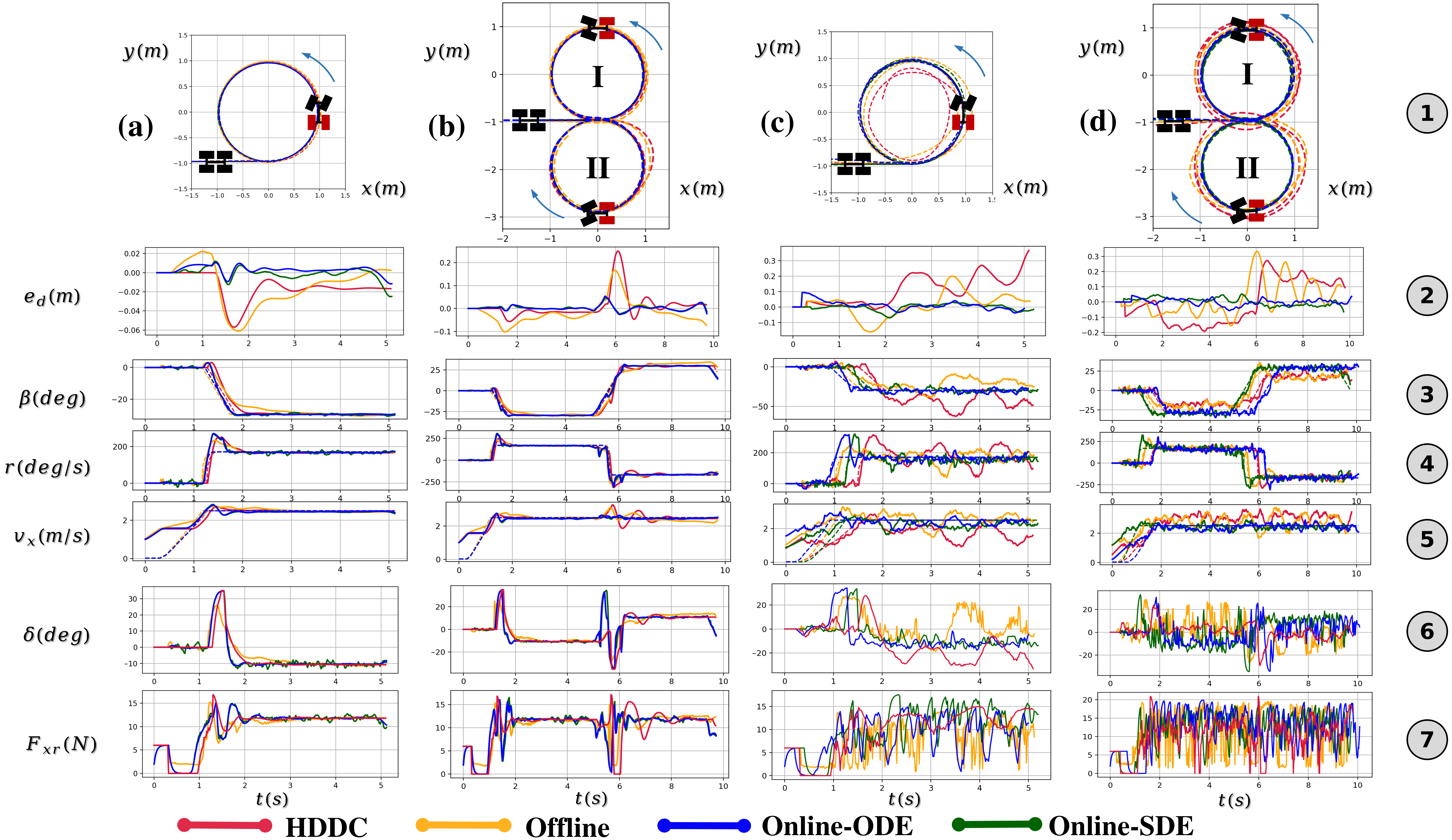}
\caption{Example Drifting Trajectories. The videos of the experiments are available at
\url{https://figshare.com/s/c62b103ac20a4962fcba}.}
\label{fig:big_arch}
\end{figure*}

\subsubsection{Testing Scenarios}
There are five cumulative test scenarios. \textbf{S1}: The nominal model; \textbf{S2}: With initial-state perturbations of up to $20\%$ of the equilibrium values; \textbf{S3}: S2 plus vehicle parameters randomized between $80\%$ and $120\%$ of their nominal values; \textbf{S4}: S3 plus integration- and force-layer noise; \textbf{S5}: S4 plus parameter drift. 

\subsubsection{Training Procedure}
We consider circular drift and eight-shaped drift. For each task, we used the HDDC controller to collect 200 episodes of expert traces. After imitation learning for 100 episodes (each with 1$k$ steps), we started to fine-tune the policy for 1200 episodes, under S1 without domain randomization. We also trained the online policy from scratch with the same condition.

We compare six controllers.
(1) \textbf{HDDC}: The expert controller;
(2) \textbf{Offline}: The policy trained via imitation learning;
(3) \textbf{Online-SDE (finetune)}: We fine-tuned the pretrained policy with our online algorithm, and used SDE to generate actions;
(4) \textbf{Online-ODE (finetune)}, which follows the same fine-tuning procedure as (3) but generates actions via ODE sampling by setting the score function to zero.

We also consider two extra controllers trained from scratch without imitation learning.
(5) \textbf{Online-SDE (scratch)}, whose policy is directly trained using our proposed online algorithm, with actions generated via SDE sampling; 
(6) \textbf{Online-ODE (scratch)}, which follows the same training procedure as (5) but generates actions via ODE sampling.

\subsubsection{Testing Procedure}
For the circular drifting task, a complete episode consists of
two stable drifting laps around the circle. For the eight-shaped
drifting task, a complete episode requires the vehicle to complete
two stable drifting laps around each of the two circles.
A trial is considered \textit{Successful} if the controller
completes the entire episode without triggering any failure condition.
A failure occurs if the magnitude of any of the error terms
$e_d$, $e_\theta$, or $e_\beta$ exceeds its corresponding threshold.
We evaluate the \textit{Success Rate} (SR) over 100 trials and use
the mean per-step reward to quantify drifting performance.

\subsection{Nominal and Disturbed Drift Experiments}
\subsubsection{Main Experiments}

Figs.~\ref{fig:circle_overall} and~\ref{fig:eight_overall} show the learning curves for circular drift and eight-shaped drift across five test scenarios. Reward curves show means with standard deviation bands. The four online training configurations outperform HDDC and Offline, with more pronounced advantages under perturbations, while HDDC generally outperforms Offline. 

\begin{table}[h]
\centering
\footnotesize
\caption{Quantitative Results in Circular Drift.}
\label{tab:circle_table}
\setlength{\tabcolsep}{2.5pt}
\renewcommand{\arraystretch}{1.05}
\begin{tabular}{@{}lrrrrrr@{}}
\toprule
& \multicolumn{3}{c}{S1 (nominal)}
& \multicolumn{3}{c}{S5} \\
\cmidrule(lr){2-4}\cmidrule(l){5-7}
Method & Ckpt. & SR (\%) & Reward & Ckpt. & SR (\%) & Reward \\
\midrule
HDDC
& --- & 100 & $-0.459$
& --- & 68 & $-1.745$ \\
Offline
& 40 & 100 & $-0.803$
& 80 & 17 & $-2.431$ \\
Online-ODE (finetune)
& 1060 & 100 & $-0.437$
& 1080 & 100 & $-0.985$ \\
Online-SDE (finetune)
& 1200 & 100 & $-0.550$
& 800 & 100 & $-1.125$ \\
Online-ODE (scratch)
& 1200 & 100 & $\mathbf{-0.374}$
& 960 & 100 & $\mathbf{-0.937}$ \\
Online-SDE (scratch)
& 1200 & 100 & $-0.543$
& 1120 & 100 & $-1.040$ \\
\bottomrule
\end{tabular}
\end{table}

\begin{table}[h]
\centering
\footnotesize
\caption{Quantitative Results in Eight-shaped Drift.}
\label{tab:eight_table}
\setlength{\tabcolsep}{2.5pt}
\renewcommand{\arraystretch}{1.05}
\begin{tabular}{@{}lrrrrrr@{}}
\toprule
& \multicolumn{3}{c}{S1 (nominal)}
& \multicolumn{3}{c}{S5} \\
\cmidrule(lr){2-4}\cmidrule(l){5-7}
Method & Ckpt. & SR (\%) & Reward & Ckpt. & SR (\%) & Reward \\
\midrule
HDDC
& --- & 100 & $-0.687$
& --- & 25 & $-1.965$ \\
Offline
& 80 & 96 & $-1.429$
& 100 & 2 & $-2.537$ \\
Online-ODE (finetune)
& 1060 & 100 & $-0.450$
& 560 & 100 & $\mathbf{-0.999}$ \\
Online-SDE (finetune)
& 1060 & 100 & $-0.608$
& 420 & 100 & $-1.167$ \\
Online-ODE (scratch)
& 1160 & 100 & $\mathbf{-0.418}$
& 360 & 100 & $-1.001$ \\
Online-SDE (scratch)
& 1000 & 100 & $-0.604$
& 680 & 100 & $-1.058$ \\
\bottomrule
\end{tabular}
\end{table}

In S1, all four online configurations achieve $100\%$ success in both tasks (Tables~\ref{tab:circle_table} and~\ref{tab:eight_table}). For circular drifting, Online-ODE (finetune) and Online-ODE (scratch) score $-0.437$ and $-0.374$, respectively, exceeding HDDC's $-0.459$. Online-SDE configurations score slightly below HDDC, which is associated with the additional noise introduced by stochastic action sampling. For eight-shaped drifting, all four online configurations outscore HDDC (Table~\ref{tab:eight_table}), with Online-ODE (scratch) reaching $-0.418$, compared with $-0.687$ for HDDC.

The advantages of online training grow as conditions become more challenging in S2--S5. The success rate for circular drifting remains close to $100\%$, while performance on eight-shaped drifting is also generally stable, with only Online-SDE (finetune) exhibiting some degradation toward the end of training. In S5 (Tables~\ref{tab:circle_table} and~\ref{tab:eight_table}), all four online configurations can achieve a $100\%$ success rate, whereas HDDC and Offline achieve success rates of $68\%$ and $17\%$ for circular drifting, and only $25\%$ and $2\%$ for eight-shaped drifting, respectively. The online configurations also maintain higher drifting scores. For example, Online-ODE (scratch) achieves $-0.937$ for circular drifting, compared with $-1.745$ for HDDC, while Online-ODE (finetune) achieves $-0.999$ for eight-shaped drifting, compared with $-1.965$ for HDDC.

The flow policy learns both tasks from scratch and converges faster than fine-tuning in eight-shaped drifting. These results indicate that, with the
proposed online method, imitation learning may not be a prerequisite in drifting. Nevertheless, when expert data are available, the
framework also supports imitation learning pretraining followed
by online fine-tuning.

Furthermore, all online configurations are trained on the nominal
model, yet maintain high success rates under increasingly severe perturbations. One possible explanation is that stochastic sampling during action generation
exposes the policy to diverse actions and the resulting state
trajectories during training, potentially providing robustness
benefits similar to those of domain randomization.

Fig.~\ref{fig:alpha_entropy} shows the evolution of the adaptive coefficient $\alpha$ and the path-wise entropy surrogate during training. Both decrease as training progresses, indicating reduced exploration. As diffusion weakens, action generation approaches ODE behavior, consistent with ODE inference after SDE training.

\subsubsection{Trajectory and State Tracking}
Figs.~\ref{fig:big_arch}(a) and (b) show representative drifting
trajectories in S1. For circular drifting, the online controllers
track the reference path more closely and achieve faster convergence
of $\beta$ and $v_x$ to their reference values. For eight-shaped
drifting, they reduce $v_x$ overshoot during direction changes,
thereby limiting the resulting lateral error $e_d$. Relative to
HDDC, they reduce the circular-drifting RMSE of $\beta$ by at least
$53\%$, to $0.010$--$0.017\,\mathrm{rad}$, and reduce the eight-shaped
drifting RMSEs of $e_d$ and $v_x$ by more than $67\%$ and at least
$17\%$, respectively.

Under the disturbances in S5, these advantages become more
pronounced, as shown in Figs.~\ref{fig:big_arch}(c) and (d).
HDDC and Offline exhibit larger path deviations and state oscillations, while the online controllers maintain
closer trajectory tracking. Their $e_d$ RMSEs are
$0.034$--$0.059\,\mathrm{m}$ for circular drifting and
$0.035$--$0.050\,\mathrm{m}$ for eight-shaped drifting,
corresponding to reductions of at least $37\%$ and $49\%$
relative to HDDC, respectively. These gains are also accompanied by improved $\beta$ and $v_x$ tracking in both tasks.

\subsection{Baseline Comparison and Ablation Study}

\subsubsection{Baseline Comparison}
We compare the flow policy with SAC~\cite{haarnoja2018soft} and PPO in the same nominal environment, allowing each method 1200 episodes of online training. For SAC and PPO, we consider both training from scratch and fine-tuning a pretrained policy. SAC is pretrained using behavior cloning (SAC-BC-transfer), while PPO is pretrained using GAIL~\cite{ho2016generative} (PPO-BC-transfer). Their randomly initialized counterparts are denoted SAC-scratch and PPO-scratch. Table~\ref{tab:training_config} summarizes the training settings.

\begin{table}[!htbp]
\centering\footnotesize
\caption{Training Configurations.}
\label{tab:training_config}
\setlength{\tabcolsep}{3pt}\renewcommand{\arraystretch}{1.05}
\begin{tabular}{@{}lccc@{}}\toprule
Setting & Flow & SAC & PPO\\\midrule
Online episodes &1200&1200&1200\\
Pretraining &Imitation&BC&GAIL\\
Actor learning rate &$3\!\times\!10^{-5}$&$3\!\times\!10^{-5}$&$3\!\times\!10^{-4}$\\
Critic learning rate &$3\!\times\!10^{-4}$&$3\!\times\!10^{-4}$&$3\!\times\!10^{-4}$\\
Critic Ensembles &2&2&2\\
Discount &0.99&0.99&0.99\\
Batch/minibatch &256&256&256\\
Replay capacity &120000&120000&---\\
Rollout length &---&---&2048\\
PPO epochs / clip &---&---&10 / 0.2\\
Integration steps &4&---&---\\\bottomrule
\end{tabular}
\end{table}

\begin{figure}[!t]
\centering
\includegraphics[width=.88\columnwidth]{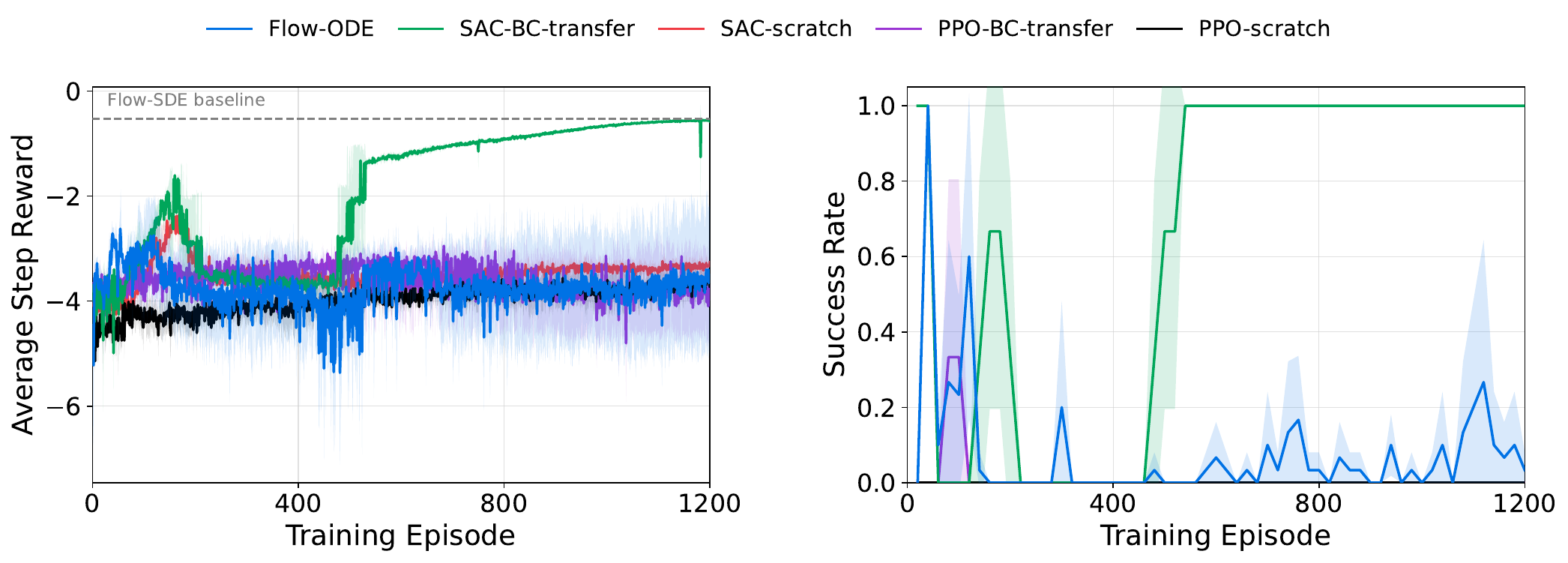}
\caption{Circular Drift Comparisons}
\label{fig:sac_compare_circle}
\vspace{1pt}
\includegraphics[width=.88\columnwidth]{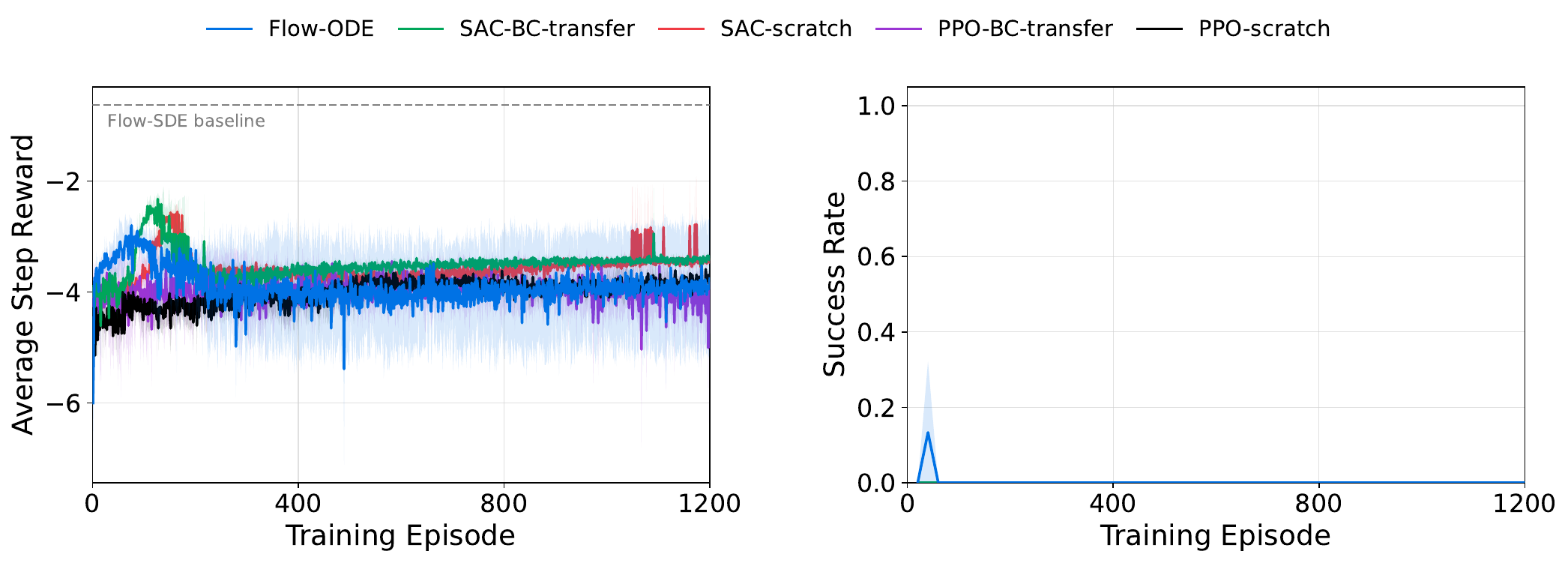}
\caption{Eight-shaped Drift Comparisons}
\label{fig:sac_compare_eight}
\vspace{5pt}
\includegraphics[width=.95\columnwidth]{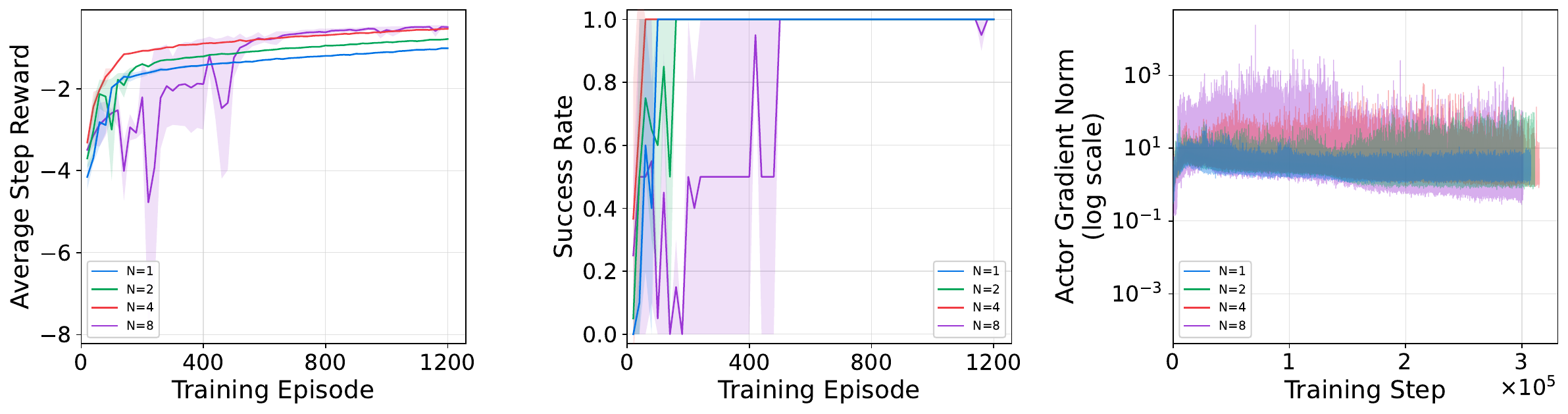}
\caption{Multiple Integration Steps in Circular Drift}
\label{fig:integration_circle}
\vspace{1pt}
\includegraphics[width=.95\columnwidth]{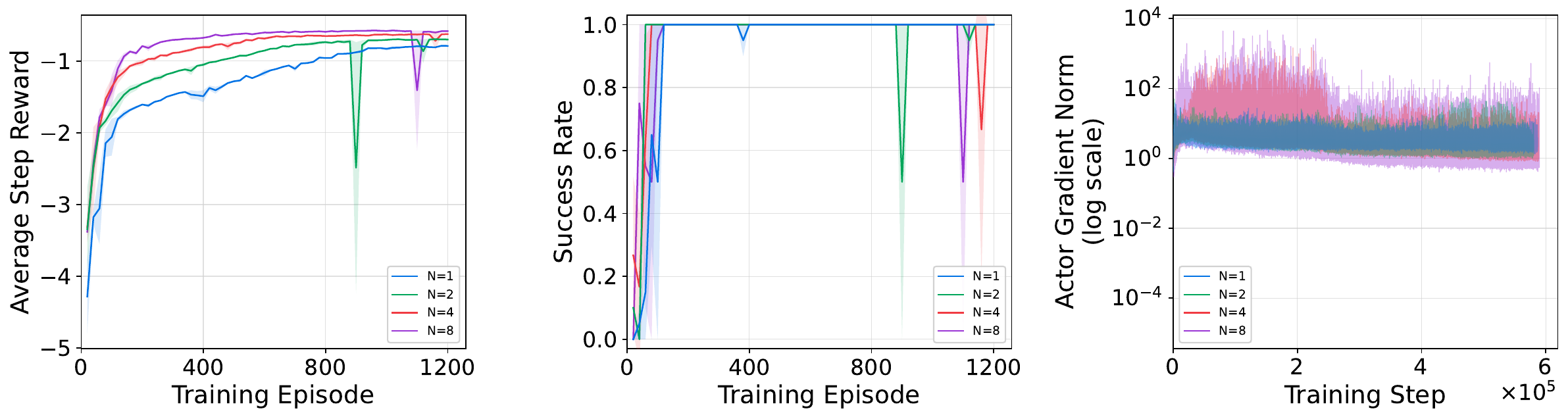}
\caption{Multiple Integration Steps in Eight-shaped Drift}
\label{fig:integration_eight}
\end{figure}

To assess the contribution of SDE-based sampling method,
we train a flow policy with ODE-based action generation,
denoted by Flow-ODE. In contrast, Online-ODE uses SDE-based action
generation during training and switches to ODE-based generation
only at test time. Thus, Flow-ODE eliminates SDE sampling from
the entire training process.

Figs.~\ref{fig:sac_compare_circle} and~\ref{fig:sac_compare_eight}
show that Flow-ODE initially improves on circular drifting but does
not sustain this improvement, and fails to learn eight-shaped
drifting. These results suggest that SDE-based sampling during
training contributes to reliable learning of both tasks.

Expert pretraining allows SAC-BC-transfer to approach the Flow-SDE baseline in circular drifting, but not to learn eight-shaped drifting.
Without pretraining, SAC records no successful evaluations in either task, and neither PPO variant learns sustained drifting within the training budget.

\subsubsection{Number of Integration Steps}

Figs.~\ref{fig:integration_circle} and~\ref{fig:integration_eight} compare flow policies using $N=1,2,4,$ and $8$ integration steps to generate each action. All settings achieve high success rates, and using more steps generally improves the reward. However, the improvement from $N=4$ to $N=8$ is modest: the final mean reward improves from $-0.539$ to $-0.500$ for circular drifting and from $-0.625$ to $-0.584$ for figure-eight drifting. More integration steps also increase the computational cost of backpropagation and result in larger actor gradients. Gradient clipping prevented gradient explosion in our experiments, but training with a large $N$ may still be more difficult to stabilize.

\begin{figure}[!htbp]
\centering
\includegraphics[width=.88\columnwidth]{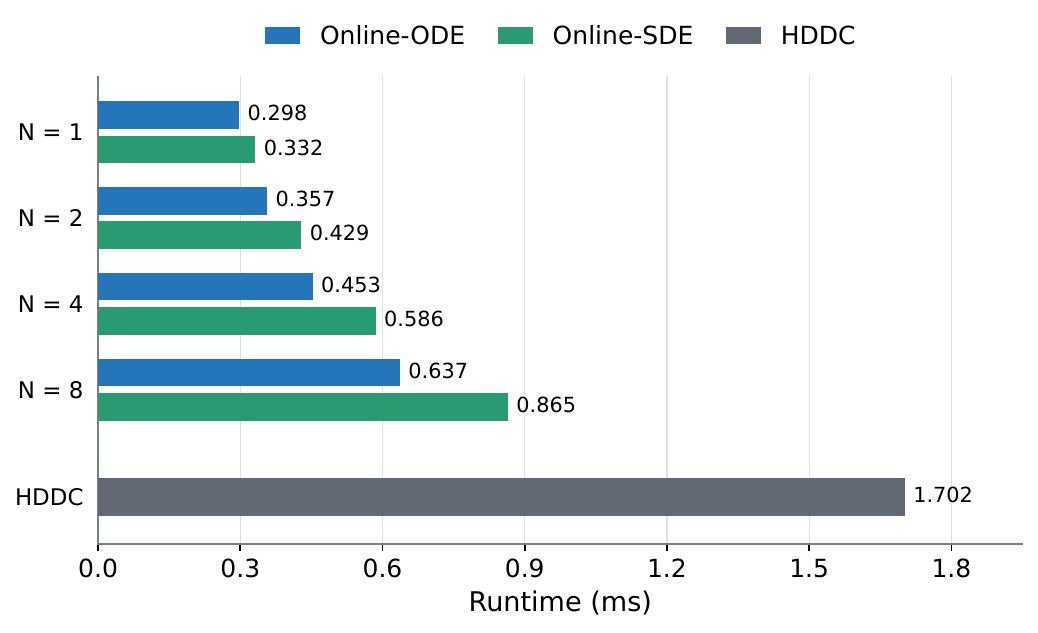}
\caption{Runtime Comparison}
\label{fig:runtime_comparison}
\end{figure}

Lastly, our flow policy supports fast inference. Fig.~\ref{fig:runtime_comparison} compares average inference times on an Intel NUC13ANKi7. All tested online configurations require less than $1\,\mathrm{ms}$ per action, compared with $1.702\,\mathrm{ms}$ for HDDC. With a fixed network architecture and number of integration steps, the predictable computational cost facilitates real-time deployment.

\FloatBarrier

\section{Conclusion}
\label{sec:conclusion}
We propose a generative framework for autonomous drifting. The framework supports both training from scratch and fine-tuning from an expert-pretrained policy. On an F1TENTH digital twin, it achieves real-time circular and figure-eight drifting with tighter tracking, higher success rates, and better robustness than the expert and offline baselines.


\bibliographystyle{IEEEtran}
\bibliography{reference}

\end{document}